\def\year{2018}\relax
\documentclass[letterpaper]{article} 
\usepackage{aaai18}  
\usepackage{times}  
\usepackage{helvet}  
\usepackage{courier}  
\usepackage{url}  
\usepackage{graphicx}  

\usepackage{booktabs}       
\usepackage{amsfonts}       
\usepackage{nicefrac}       
\usepackage{microtype}      

\usepackage{comment}		
\usepackage{adjustbox}		
\usepackage{calc}			
\usepackage{wrapfig}

\usepackage{amsmath}
\usepackage{amssymb}
\usepackage{amsthm}
\usepackage{multirow}
\usepackage[hang,flushmargin]{footmisc}
\usepackage[neverdecrease]{paralist}
\usepackage{subcaption}
\usepackage{algorithm}
\usepackage{algorithmic}
\usepackage{calc}
\usepackage{color}

\usepackage{tikz}

\newcommand\tcircled[1]{\raisebox{.5pt}{\textcircled{\raisebox{-.95pt} {#1}}}}

\newcommand{\bT}{\mathbf{T}}

\newcommand{\bX}{\mathbf{X}}

\newcommand{\btheta}{\mbox{\boldmath $\theta$}}

\newcommand{\be}{\begin{eqnarray}}
\newcommand{\ee}{\end{eqnarray}}
\newcommand{\bee}{\begin{eqnarray*}}
\newcommand{\eee}{\end{eqnarray*}}

\newcommand{\matrixb}{\left[ \begin{array}}
\newcommand{\matrixe}{\end{array} \right]}

\usepackage{xspace}
\makeatletter
\def\onedot{\ifx\@let@token.\else.\null\fi\xspace}
\makeatother

\def\eg{\textit{e.g}\onedot} 
\def\ie{\textit{i.e}\onedot}

\def\wrt{\textit{w.r.t}\onedot} 
\def\etal{\textit{et al}\onedot}

\newcommand{\Tref}[1]{Table~\ref{#1}}
\newcommand{\Eref}[1]{Eq.~(\ref{#1})}
\newcommand{\Fref}[1]{Fig.~\ref{#1}}

\newcommand{\Cref}[1]{Chap.~\ref{#1}}

\newcommand{\R}{\mathbb{R}}

\begin{document}
%
\title{Co-domain Embedding using Deep Quadruplet Networks\\for Unseen Traffic Sign Recognition}
\author{Junsik Kim\qquad Seokju Lee\qquad Tae-Hyun Oh${}^\dagger$\qquad In So Kweon\\
Dept. of Electrical Engineering, KAIST, Daejeon, Korea\\
${}^\dagger$MIT CSAIL, Cambridge, US\\
\texttt{\{mibastro,seokju91\}@gmail.com, taehyun@csail.mit.edu${}^\dagger$, iskweon@kaist.ac.kr}\\
}
\maketitle
\begin{abstract}
Recent advances in visual recognition show overarching success by virtue of large amounts of supervised data. 
However, the acquisition of a large supervised dataset is often challenging. This is also true for intelligent transportation applications, i.e., traffic sign recognition. For example, a model trained with data of one country may not be easily generalized to another country without much data.
We propose a novel feature embedding scheme for unseen class classification when the representative class template is given. Traffic signs, unlike other objects, have official images. We perform co-domain embedding using a quadruple relationship from real and synthetic domains. Our quadruplet network fully utilizes the explicit pairwise similarity relationships among samples from different domains. We validate our method on three datasets with two experiments involving one-shot classification and feature generalization. The results show that the proposed method outperforms competing approaches on both seen and unseen classes.
\end{abstract}

\section{Introduction}

Recent advances in the field of computer vision have provided highly cost-effective solutions for developing advanced driver assistance systems (ADAS) for automobiles.
Furthermore, computer vision components are becoming indispensable to improve safety and to achieve AI in the form of fully automated, self-driving cars.
This is mostly by virtue of the success of deep learning, which is regarded to be due to the presence of large-scale supervised data, proper computation power and algorithmic advances~\cite{book_deeplearn}. 

Among all ADAS related problems, in this paper, we tackle unseen traffic sign recognition.
A distinctive difference related to this problem as regards traditional recognition problems is that synthetic traffic-sign templates are exploited as representative anchors, whereby classification can be done for an actual query image by finding the minimum distance to the templates of each class (\ie, few-shot learning with domain difference).


In reality, traffic signs differ depending on the country, but one may obtain synthetic templates from traffic-related public agencies.
Nonetheless, the diversity of templates for a single class is limited; hence, we focus on scenarios of challenging one-shot classification~\cite{kochsiamese,lake2015human,miller2000learning} with domain adaptation, where a machine learning model must generalize to new classes not seen in the training phase given only a few examples of each of these classes but from different domains.

In practice, this type of model is especially useful for ADAS in that: 1) one can avoid high cost re-training from scratch, 2) one can avoid annotating large-scale supervised data, and 3) it is readily possible to adapt the model to other environments.

Given the success of deep learning, a naive approach for the few-shot problem would be to re-train a deep learning model on a new scarce dataset. However, in this limited data regime, this type of naive method will not work well, likely due to severe over-fitting~\cite{lake2015human}.
While people have an inherent ability to generalize from only a single example with a high degree of accuracy, the problem is quite difficult for machine learning models~\cite{lake2015human}.

\begin{figure*}
	\centering	
	{\includegraphics[width=1\linewidth]{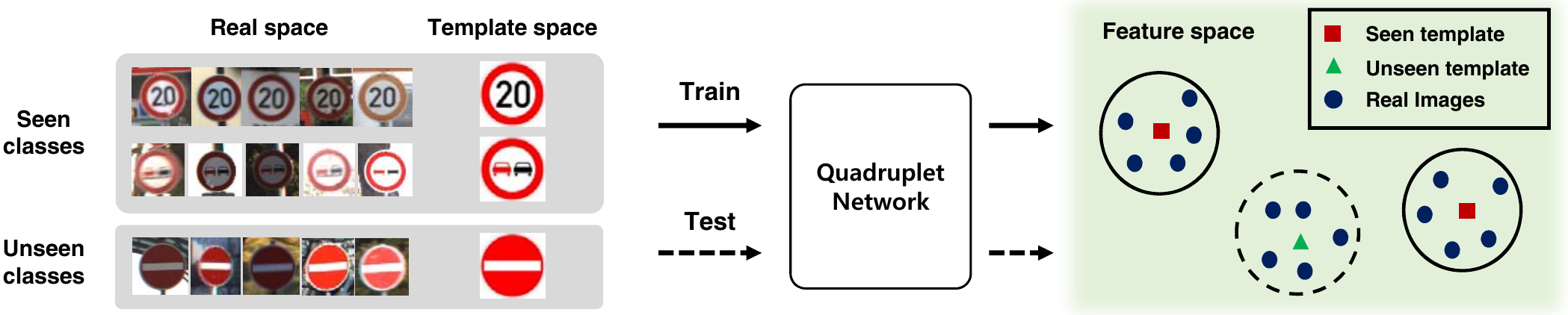}}\hspace{2mm}
	\caption{Illustration of a synthetic template and real image co-domain mapping.}
	\label{fig:idea}
\end{figure*}

Thus, our approach is based on the following hypotheses: 1) the existence of a co-embedding space for synthetic and real data, and 2) the existence of an embedding space where real data is condensed around a synthetic anchor for each class. We illustrate the idea in \Fref{fig:idea}.
Taking these into account, we learn two non-linear mappings using a neural network. The first involves mapping for a real sample into an embedding space, and the second involves mapping of a synthetic anchor onto the same metric space.
We leverage the quadruplet relationship to learn non-linear mappings, which can provide rich information to learn generalized and discriminative embeddings.
Classification is then performed for an embedded query point by simply finding the nearest class anchor. 
Despite its simplicity, our method outperforms with a margin in the unseen data regime.

\section{Related work}
Our problem can be summarized as a modified one-shot learning problem that involves heterogeneous domain datasets.
This type of problem has gained little attention. Furthermore, to the best of our knowledge, our work is the first to tackle unseen traffic sign recognition with heterogeneous domain data;
we therefore summarize the work most relevant to our proposed method in this section.

Non-parametric models such as nearest neighbors are useful in few-shot classification~\cite{vinyals2016matching,santoro2016meta}, in that they can be naturally adapted to new data and do not require training of the model. However, the performance depends on the chosen metric~\cite{atkeson1997locally}.\footnote{For up-to-date thorough surveys on metric learning, please refer to \cite{kulis2013metric,bellet2015metric}.}
To overcome this, 
Goldberger~\etal\cite{goldberger2004neighbourhood} propose  the neighborhood components analysis (NCA) and that learns the Mahalanobis distance to maximize the accuracy of the $K$-nearest-neighbor ($K$-NN).
Weinberger~\etal\cite{weinberger2009distance}, who studied large margin nearest neighbor (LMNN) classification, also maximize the $K$-NN accuracy with a hinge loss that encourages the local neighborhood of a point to contain other points with identical labels with some margin, and vice versa. 
Our work also adopts the hinge loss with a margin in the same spirit of Weinberger~\etal
Because NCA and LMNN are limited on linear model, each is extended to a non-linear model using a deep neural network in \cite{salakhutdinov2007learning} and \cite{min2009deep} respectively.

Our work may be regarded as a non-linear and quadruple extension of 
Mensink~\etal\cite{mensink2013distance} and
Perrot~\etal\cite{perrot2015regressive} to one-shot learning, 
in that they leverage representative auxiliary points for each class instead of individual examples of each class.
These approaches are developed to adapt to new classes rapidly without re-training; however, they are designed to handle cases where novel classes come with a large number of samples. 
In contrast, in our scenario, a representative example is given explicitly as a template image of a traffic sign.
Given such a template, we learn non-linear embedding in an end-to-end manner without pre-processing necessary in both Mensink~\etal and Perrot~\etal to obtain the representatives.

All of these NN classification schemes learn the metric via a pairwise relationship.
In the recent metric learning literature, there have been attempts~\cite{wang2014learning,hoffer2015deep,law2016learning,chen2017beyond,huang2016learning} to go beyond learning metrics using only a pairwise relationship (\ie, 2-tuple, \eg, Siamese~\cite{bromley1993signature,chopra2005learning,hadsell2006dimensionality}): triplet~\cite{weinberger2009distance,wang2014learning,hoffer2015deep}, quadruplet~\cite{law2016learning} and quintuplet~\cite{huang2016learning}.
The use of tuples of more than a triple relationship may be inspired from the argument of Kendall and Gibbons~\cite{kendall1990rank}, who argued that humans are better at providing relative (\ie, at least triplet-wise) comparisons than absolute comparisons (\ie, pairwise).  
While our method also exploits a quadruple relationship, the motivation behind this composition is rather specific for our problem definition, in which two samples from template sets and two samples from real sets have clear combinatorial pairwise relationships. More details will be described later.

Other one-shot learning approaches take wholly different notions.
Koch~\etal\cite{kochsiamese} uses Siamese network to classify whether two images belong to the same class.
To address one-shot learning for character recognition, Late~\etal\cite{lake2015human} devise a hierarchical Bayesian generative model with knowledge of how a hand written character is created.
A recent surge of models, such as a neural Turing machine~\cite{graves2014neural}, stimulate the meta-learning paradigm~\cite{santoro2016meta,vinyals2016matching,ravi2017fewshot} for few-shot learning.
Comparing to these works that have limited memory capacities, 
the NN classifier has an unlimited memory and can, automatically store and retrieve all previously seen examples. 
Furthermore, in the few-shot scenario, the amount of data is very small to the extent that a simple inductive bias appears to work well without the need to learn complex input-sensitive embedding~\cite{vinyals2016matching,santoro2016meta,ravi2017fewshot}, as we do so. 
This provides the $k$-NN with a distinct advantage.


Moreover, because we have two data sources, synthetic templates and real examples, a domain difference is naturally introduced in our problem.
There is a large amount of research that smartly solves domain adaptation (refer to the survey by Csurka~\etal\cite{book_DA} for a thorough review), but we deal with this by simply decoupling the network parameters from each other, the template and the real domains.
This method is simple, but in the end it generalizes well owing to the richer back-propagation gradients from the quadruple combinations.

\section{Quadruple network for jointly adapting domain and learning embedding}

Our goal is to learn embeddings, such that different domain examples are embedded into a common metric space and where their embedded features favor to be generalized as well as discriminative.

To this end, we leverage a quadruplet relationship, consisting of two anchors of different classes and two others for examples corresponding to the anchors. We first describe the quadruple (4-tuple) construction in the following section. Subsequently, we define the embeddings followed by the 
objective functions and quadruplet network.

\noindent\textbf{Notation}\quad
We consider two imagery datasets: the template set $\mathcal{T} {=} \{ (\bT, y) \}$, where each $\bT$ denotes a representative template image and $y{\in}\{1, \dots, C\}$ is the corresponding label (out of the $C$ class), and the real example set $\mathcal{X} {=} \{ \mathcal{X}_k \}_{k=1}^C$, where $\mathcal{X}_k$ is the set of real images $\{\bX\}$ of class-$k$.
For simplicity, we use $\bT_k$ ($\bX_k$) to denote a sample labeled with class-$k$.

We define Euclidean embeddings as $f(\cdot)$, where $f(\mathbf{x})$ maps a high-dimensional vector $\mathbf{x}$ into a $D$-dimensional feature space, \ie, $\mathbf{e}  {=} f(\mathbf{x}) \in \R^D$.

\subsection{Quadruple (4-tuple) construction} 
Our idea is to embed template and real domain data into the same feature space such that a template sample acts as an anchor point on the feature space and real samples relevant to the anchor form a cluster around it, as illustrated in \Fref{fig:idea}. 
Specifically, we aim to achieve two properties for an embedded feature space: 1) distinctiveness between anchors is favored, and 2) real samples must be mapped close to the anchor that corresponds to the same class. 

To leverage the relational information, we define a quadruple, a basic element, by packing two template images from different classes and two real images corresponding to respective template classes, \ie, for simplicity, two classes $A$ and $B$ are considered, then $(\bT_A, \bT_B, \bX_A, \bX_B)$.
From pairwise combinations within the quadruple, we can reveal several types of relational information as follows:\\[3pt]
\noindent
\tcircled{1} $\bT_A$ should be far from $\bT_B$ in an embedding space, \\
\noindent
\tcircled{2} $\bX_A$ should be far from $\bX_B$ in an embedding space, \\
\noindent
\tcircled{3} $\bX_A$ (or $\bX_B$) should be close to $\bT_A$ (or $\bT_B$) in an embedding space,\\
\noindent
\tcircled{4} $\bT_A$ (or $\bT_B$) should be far from $\bX_B$ (or $\bX_A$) in an embedding space,\\  
whereby we derive the final objective function.
These relations depicted in \Fref{fig:quadruplet_relation}.

\paragraph{Quadruple sampling}
We sampled two templates $(\bT_{A}, y)$ and $(\bT_{B}, y')$ from template set $\mathcal{T}$ while guaranteeing two different classes, followed by the two real images of $\bX_{A} \in \mathcal{X}_{y}$ and $\bX_{B} \in \mathcal{X}_{y'}$.

\paragraph{Comparison to other tuple approaches}
In metric learning, the most common approaches would be the Siamese~\cite{bromley1993signature,chopra2005learning,hadsell2006dimensionality} and triplet~\cite{weinberger2009distance,wang2014learning,hoffer2015deep}, which typically use 2- and 3-tuples, respectively. 
From the given tuple, they optimize with the pairwise differences. This concept can be viewed as follows: given a tuple, the Siamese has only a single source of loss (and its gradient), while triplets utilize two sources, \ie, (query, positive) and (query, negative).
With this type of simple comparison, we can intuitively conjecture higher stability or performance of triplet network over Siamese network.

Law~\etal\cite{law2016learning} deal with a particular ambiguous quadruple relationship $(\mathsf{a}) {\prec}(\mathsf{b}){\simeq}(\mathsf{c}){\prec}(\mathsf{d})$ by forcing the difference between $(\mathsf{b})$ and $(\mathsf{c})$ to be smaller than the difference between $(\mathsf{a})$ and $(\mathsf{d})$.
Huang~\etal\cite{huang2016learning} (quintuplet, 5-tuple) is devised a means to handle class imbalance issues by leveraging the relationships among three levels (\eg, strong, weaker and weakest in terms of a cluster analysis) of positives and negatives.
Comparing the motivations of these approaches, for instance  indefinite relativeness, our quadruple comes from a specific relationship, \ie, heterogeneous domain data.
Moreover, it is important to note that our quadruple provides richer information (a total of $6$ pairwise information) compared to the method of Law~\etal (quadruplet) which leverages a single constraint from a quadruple. It is even richer than Huang~\etal (quintuplet), who provides $3$ constraints.

\subsection{Quadruplet Network}
\label{sec:quad_loss}

Given the defined quadruple, we propose a quadruple metric learning that learns to embed template images and real images into a common metric space, say $\R^D$, through an embedding function.
In order to deal with non-linear mapping, the embedding $f$ is modeled as a neural network, of which set of weight parameters are denoted as $\btheta$.
Since we deal with data from two different domains, template and real images, we simply use two different neural networks $\btheta_\mathsf{T}$ and $\btheta_\mathsf{R}$ for the template and real images respectively, expressed as $f_\mathsf{T}(\cdot) {=} f(\cdot\,; \btheta_\mathsf{T})$ and $f_\mathsf{R}(\cdot) {=} f(\cdot\,; \btheta_\mathsf{R})$, such that we can adapt both domains. Now, we are ready to define the proposed quadruple network.

\begin{figure}[t]
   \centering
   \captionsetup[subfigure]{aboveskip=4pt}
   \begin{tabular}{c@{\hspace{0mm}}c@{\hspace{0mm}}}
      \subcaptionbox{\label{fig:quadruplet_structure}}{\includegraphics[width=0.95\linewidth]{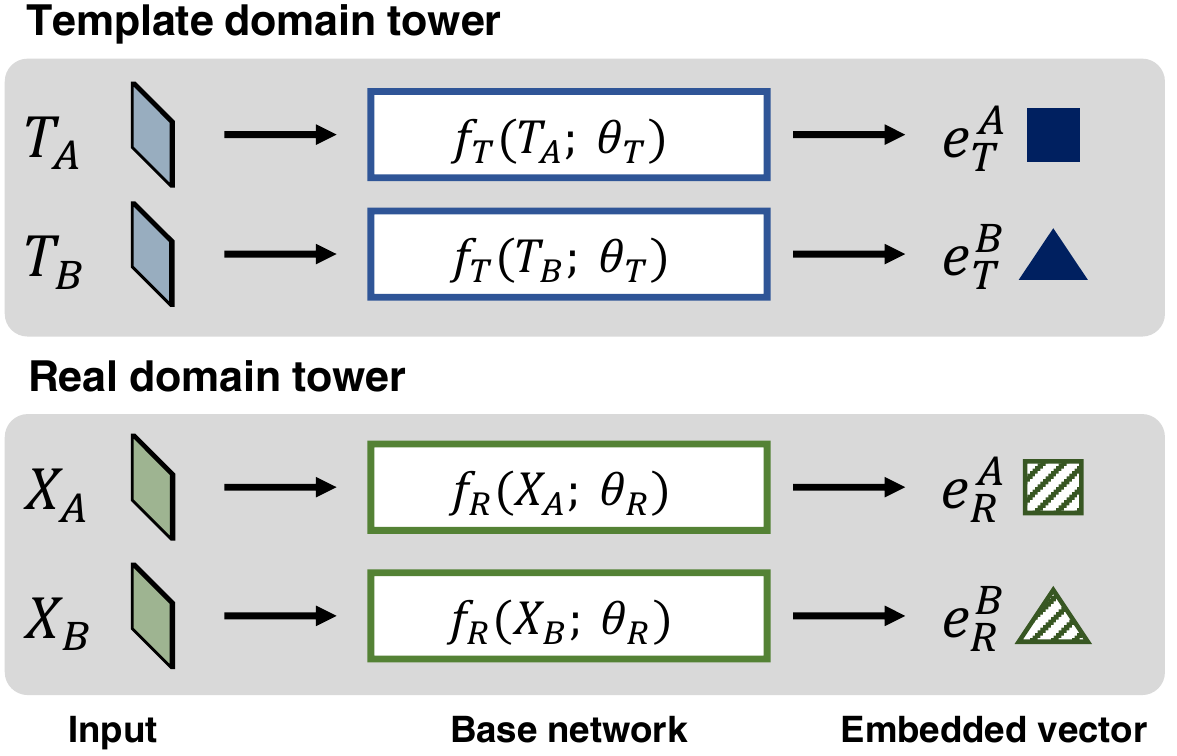}}
      \\
      \subcaptionbox{\label{fig:quadruplet_relation}}{\includegraphics[width=0.7\linewidth]{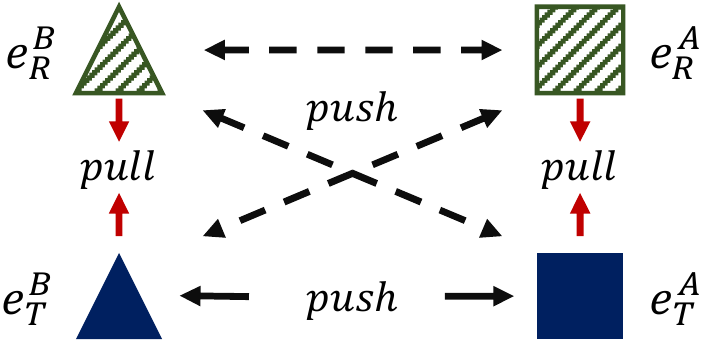}}          
   \end{tabular}
   \caption{(a) Quadruplet network structure. (b) pairwise relations of embedded vectors.}
   \label{fig:QuadNet}
\end{figure}

The proposed quadruple network $Q$ is composed of two Siamese networks, the weights of which are shared within each pair. 
One part embeds features from template images and the other part for real images. 
Quadruple images from each domain are fed into the corresponding Siamese networks (depicted in \Fref{fig:quadruplet_structure}), and denoted as
\begin{equation}
\begin{aligned}
&Q\big( \left( \bT_A,\bT_B,\bX_A,\bX_B \right); \btheta_\mathsf{T}, \btheta_\mathsf{R} \big) \\
&=\big[ f_\mathsf{T}(\bT_A), f_\mathsf{T}(\bT_B), f_\mathsf{R}(\bX_A), f_\mathsf{R}(\bX_B) \big] \\
&= \big[ \mathbf{e}_\mathsf{T}^A, \mathbf{e}_\mathsf{T}^B, \mathbf{e}_\mathsf{X}^A, \mathbf{e}_\mathsf{X}^B \big],
\end{aligned}
\label{eqn:quad}
\end{equation}
for two different arbitrary classes $A$ and $B$,
where $\mathbf{e} {\in} \R^d$ represents the embedded vector mapped by the embedding function $f$.

\paragraph{Loss function}

We mainly utilize the $l_2$ hinge embedding loss with a margin to train the proposed network. 
Given outputs quadruplet features $\big[ \mathbf{e}_\mathsf{T}^A, \mathbf{e}_\mathsf{T}^B, \mathbf{e}_\mathsf{R}^A, \mathbf{e}_\mathsf{R}^B \big]$, 
we have up to six pairwise relationships, as shown in \Fref{fig:quadruplet_relation}), and obtain six pairwise Euclidean feature distances as $\{d(\mathbf{e}_\mathsf{j}^k, \mathbf{e}_\mathsf{j'}^{k'})\}_{j,j' \in \{\mathsf{T},\mathsf{R}\}}^{k,k' \in \{A,B\}}$, where $d(\mathsf{a},\mathsf{b}) = \|\mathsf{a}{-}\mathsf{b}\|_2$ denotes the Euclidean distance.
Let $h_m(d)$ denote the hinge loss with margin $m$ as $h_m(d) = \max(0,m{-}d)$, where $d$ is the distance value. 
We then encourage the embedded vectors with the same label pairs (\ie, if $k = k'$) to be close by applying the loss $h_{-m}(-d)$, while pushing away the different label pairs by applying $h_{m'}(d)$.

To induce the embedded feature space with the aforementioned two properties in the \emph{Quadruple Construction} section, the minimum number of necessary losses is three, while we can exploit up to six losses; hence, we have several choices to formulate the final loss function, as described below:

\begin{eqnarray}
&\hspace{-4mm}\textsf{HingeM-3}:  {L3_{m,m'}} {=}&\hspace{-2mm} h_m\left( d\left(\mathbf{e}_\mathsf{T}^A, \mathbf{e}_\mathsf{T}^{B} \right) \right) \nonumber\\
&&
\hspace{-24mm}{+}
{h_{-m'}}\left( d\left(\mathbf{e}_\mathsf{T}^A, \mathbf{e}_\mathsf{X}^{A} \right) \right)
+
h_{-m'}\left( d\left(\mathbf{e}_\mathsf{T}^B, \mathbf{e}_\mathsf{X}^{B} \right) \right),
\label{eqn:loss3}\\
&\hspace{-4mm}\textsf{HingeM-5}:  {L5}_{m,m'} {=}&\hspace{-2mm} {L3}_{m,m'} + h_{m}\left( d\left(\mathbf{e}_\mathsf{T}^A, \mathbf{e}_\mathsf{X}^{B} \right) \right) \nonumber\\
&&+
h_{m}\left( d\left(\mathbf{e}_\mathsf{X}^A, \mathbf{e}_\mathsf{T}^{B} \right) \right),
\label{eqn:loss5}\\
&\hspace{-4mm}\textsf{HingeM-6}:  {L6}_{m,m'} {=}&\hspace{-2mm} {L5}_{m,m'} + h_{m}\left( d\left(\mathbf{e}_\mathsf{X}^A, \mathbf{e}_\mathsf{X}^{B} \right) \right).
\label{eqn:loss6}
\end{eqnarray}

In addition, inspired by the contrastive loss~\cite{chopra2005learning}, we also adopt  an alternative loss by replacing the $h_{-m}(-d)$ terms in \Eref{eqn:loss3} with directly minimizing $d$.\footnote{This can be viewed as a $l_1$ version of the contrastive loss~\cite{chopra2005learning}; this is how  \texttt{HingeEmbeddingCriterion} is implemented for the pairwise loss in \texttt{Torch7}~\cite{collobert2011torch7}.}
We denote this alternative loss simply as \textsf{contrastive} with a slight abuse of the terminology.
We will compare these losses in the \emph{Experiments} section. Analogous to traditional deep metric learning, training with these losses can be done with a simple SGD based method.
By using shared parameter networks, the back-propagation algorithm updates the models \wrt several relationships; \eg, in the \textsf{HingeM-6} case, the template tower is updated \wrt 5 pairwise relationships (\ie, $(\mathsf{T}^A,  \mathsf{T}^B), (\mathsf{T}^A,  \mathsf{X}^A),
(\mathsf{T}^B,  \mathsf{X}^B), (\mathsf{T}^A,  \mathsf{X}^B), (\mathsf{X}^A,  \mathsf{T}^B)$), analogously for the real tower.

\section{Experiments}

\subsection{Experiment setup}

\paragraph{Competing methods}
We compare the proposed method with other deep neural network based previous works and the additionally devised baseline, as follows:
{
	\setdefaultleftmargin{4mm}{}{}{}{}{}
	\begin{itemize}
		\item[-] \emph{IdsiaNet}~\cite{cirecsan2012multi} is a competition winner of the German Traffic-Sign Recognition Benchmark~\cite{stallkamp2012man} (GTSRB). We directly used an improved implementation~\cite{impidsia:repository}.\footnote{The improved performance is reported in \cite{impidsia:repository} with an even simpler architecture and without using ensemble. For brevity, we denote this improved version simply as \emph{IdsiaNet}. Detailed information can also be found in the supplementary material.}
		For all of the experiments, IdsiaNet is exhaustively compared as a supervised model baseline, in the same philosophy of the deep generic feature~\cite{donahue2014decaf}. Moreover, templates are not used for training.
		For a fair comparison, the architecture itself is adopted as the base network for the following models.
		\item[-] \emph{Triplet}~\cite{hoffer2015deep} (Hoffer~\etal) is similar to our model but with triplet data.
		Three weight shared networks are used, while in training, we randomly sampled triplets within the real image set with labels, such that no template image is exposed during triplet training. This training method is consistent with Hoffer~\etal
		
		\item[-] \emph{Triplet-DA} (domain adaptation) is a variant devised by us to test our hypothesis that involving different domain templates as an anchor is beneficial. 
		Three weight shared networks are used, and for triplet sampling, we sample one template $(\bT, y)$ from template set $\mathcal{T}$ and then sample positive and negative real images from $\mathcal{X}_y$ and $\mathcal{X}_{k \neq y}$ respectively.
	\end{itemize}
}

\paragraph{Implementation details} For fairness, all of the details are equally applied unless otherwise specified. All input images are resized to $48{\times} 48$ and the mean intensity of training set is subtracted. 
We did not perform any further preprocessing, data augmentation, or ensemble approach. 

We use the same improved IdsiaNet~\cite{impidsia:repository} for the base network of \emph{Triplet}, \emph{Triplet-DA} and our \emph{Quadruplet}, but replace the output dimension of the final layer \texttt{FC2} to be $\R^D$ without a softmax layer. Every model is trained from scratch. Most of the hyper parameters are based on the implementation~\cite{impidsia:repository} with a slight modification (\ie, fixed learning rates: $10^{-3}$, momentum: $0.9$, weight decay: $10^{-4}$, mini-batch size: 100, optimizer: stochastic gradient descent). 
Models are trained until convergence is reached, \ie, $\tfrac{L_t - L_{t-1}}{L_{t-1}} {<} 5\%$ where $L_t$ denotes the loss value at $t$-th iteration. We observed that the models typically converged at around $15k$-$20k$ iterations.
All networks were implemented using \texttt{Torch7}~\cite{collobert2011torch7}.

\subsection{Dataset}

\begin{wrapfigure}[9]{R}{0.23\textwidth}
\centering
\includegraphics[width=0.9\linewidth]{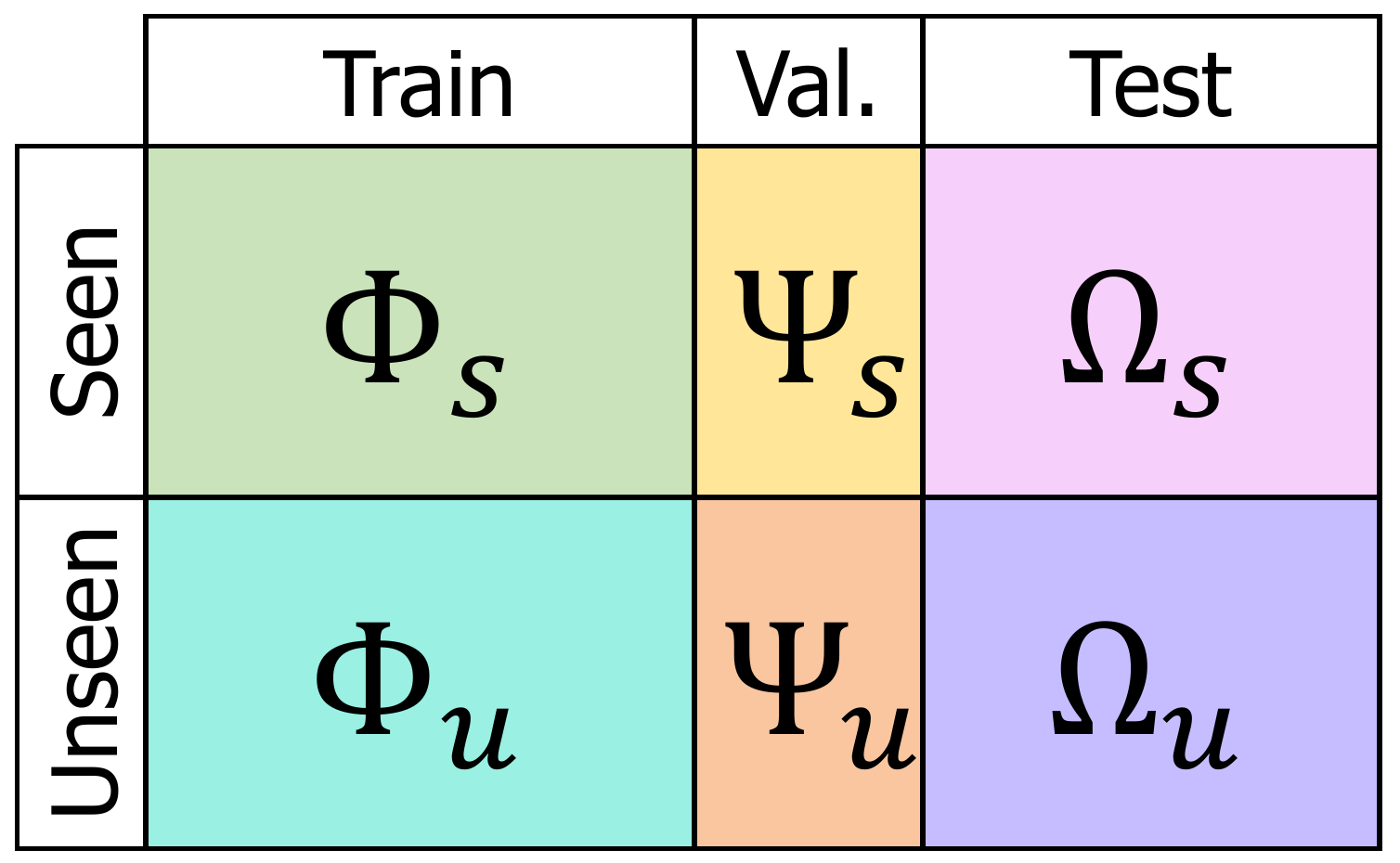}
\caption{Partitions of dataset.}
\label{fig:datapartition}
\end{wrapfigure}

We use two traffic-sign datasets, GTSRB~\cite{stallkamp2012man} and Tsinghua-Tencent 100K~\cite{zhu2016traffic} (TT100K). 
Since our motivation can be considered to involve dealing with a deficient data regime and a data imbalance caused by rare classes, we additionally introduce a subset split from GTSRB, referred to here as GTSRB-sub. 
To utilize the dataset properly for our evaluation schemes, given the train and test set splits provided by the authors, we further split them into seen, unseen and validation partitions, as illustrated in \Fref{fig:datapartition}.
The validation set is constructed by random sampling from the given training set.
A description of the dataset construction process follows.
{For more details,} the reader can refer to the supplementary material.

\noindent\textbf{GTSRB-all}\quad
GTSRB contains 43 classes. The dataset contains severe illumination variation, blur, partial shadings and low-resolution images. The benchmark provides partitions into {training} and {test} sets. The training set contains $39k$ images, where $30$ consecutively captured images are grouped, called a ``track''. The test set contains $12k$ images without continuity and, thus does not form tracks.  
We selected 21 classes that have the fewest samples as unseen classes with the remaining 22 classes as the seen classes. Template images are involved in the dataset. 

\noindent\textbf{GTSRB-sub}\quad{
	To analyze the performance of the deficient data regime, we created a subset of GTSRB-all, forming a smaller but class-balanced training set with sharing the test set of GTSRB-all. Hence, numeric results will be compatibly comparable to that from GTSRB-all. For the training set, we randomly select 7 tracks (\ie, 210 images) for each seen classes.}

\noindent\textbf{TT100K-c}\quad{ The TT100K detection dataset includes over 200 sign classes. 
	We cropped traffic sign instances from the scenes to build the classification dataset (called TT100K-c). 
	Although it contains the huge number of classes, most of  the classes do not have enough instances to conduct the experiment. 
	We only selected classes having official templates\footnote{The official templates are provided by  the Beijing Traffic Management Bureau at,
		\url{http://www.bjjtgl.gov.cn/english/traffic signs/index.html}.} available and a sufficient number of instances. 
	We split \hbox{TT100K-c} into the train and test set according to the number of instances. 
	We set 24 classes with ${\geq} 100$ instances for seen classes and select 12 unseen classes as those having $50$-$100$ instances.
	The {training} set includes half of the seen class samples, and the other half is sorted into the test set. 
}

\subsection{One-shot classification}

We perform one-shot classification by 1-nearest neighbor (1-NN) classification.
For 1-NN, the Euclidean distance is measured between embedding vectors by forward-feeding a query (real image) and the anchors (template images) to the network, after which the most similar anchor out of the $C$ classes is found ($C$-way classification).
For the NN performance, we measure the average accuracy for each class.
The seen class performance is also reported for a reference purpose.

\paragraph{Self-Evaluation}
\label{sec:selfeval}

\begin{table}[t]
	\centering
	
	\caption{Evaluations of the proposed quadruplet network with varying parameters. Evaluation is conducted on the validation set of GTSRB-all. Accuracy ($\%$) on validation set is reported.}
	\label{table:selfeval}
	
	\begin{subtable}[h]{0.3\textwidth}
		\centering
		\resizebox{!}{0.2\textwidth}{%
			\begin{tabular}{cccc}
				\toprule
				\multirow{3}{*}{\begin{tabular}[c]{@{}c@{}}Embedding dim\\( HingeM-5 ) \end{tabular}} 
				& \multicolumn{3}{c}{Top1 NN}  \\ 
				\cmidrule{2-4}
				& Avg.       & Seen        & Unseen       \\ 	\midrule
				$D=50$             & 67.1 		 & 92.2          & 40.8 	 \\ 
				$D=100$		    & 69.1 		 & 95.3          & 41.6 	 \\ 
				$D=150$           & 68.9 		 & 94.2          & 42.4 	 \\ 
				\bottomrule
			\end{tabular}
		}
		\caption{Varying embedding dimension $d$.}
		\label{tab:selfeval_dim}
	\end{subtable}
    
	\begin{subtable}[h]{0.3\textwidth}
		\centering
		\resizebox{!}{0.2\textwidth}{%
			\begin{tabular}{cccc}
				\toprule
				\multirow{2}{*}{\begin{tabular}[c]{@{}c@{}}Loss terms\\(dim 100) \end{tabular}} 
				& \multicolumn{3}{c}{Top1 NN}  \\ 
				\cmidrule{2-4}
				& Avg.       & Seen        & Unseen       \\ 	\midrule
				\textsf{HingeM}-3             & 67.3 		 & 93.1          & 40.3 	 \\ 
				\textsf{HingeM}-5 		    & 69.1 		 & 95.3          & 41.6 	 \\ 
				\textsf{HingeM}-6            & 68.9 		 & 97.3          & 39.2 	 \\ 
				\bottomrule
			\end{tabular}
		}
		\caption{Varying number of pairwise loss terms.}
		\label{tab:selfeval_loss}
	\end{subtable}
\end{table}

\begin{figure}[t]
	\centering
\includegraphics[width=0.7\linewidth]{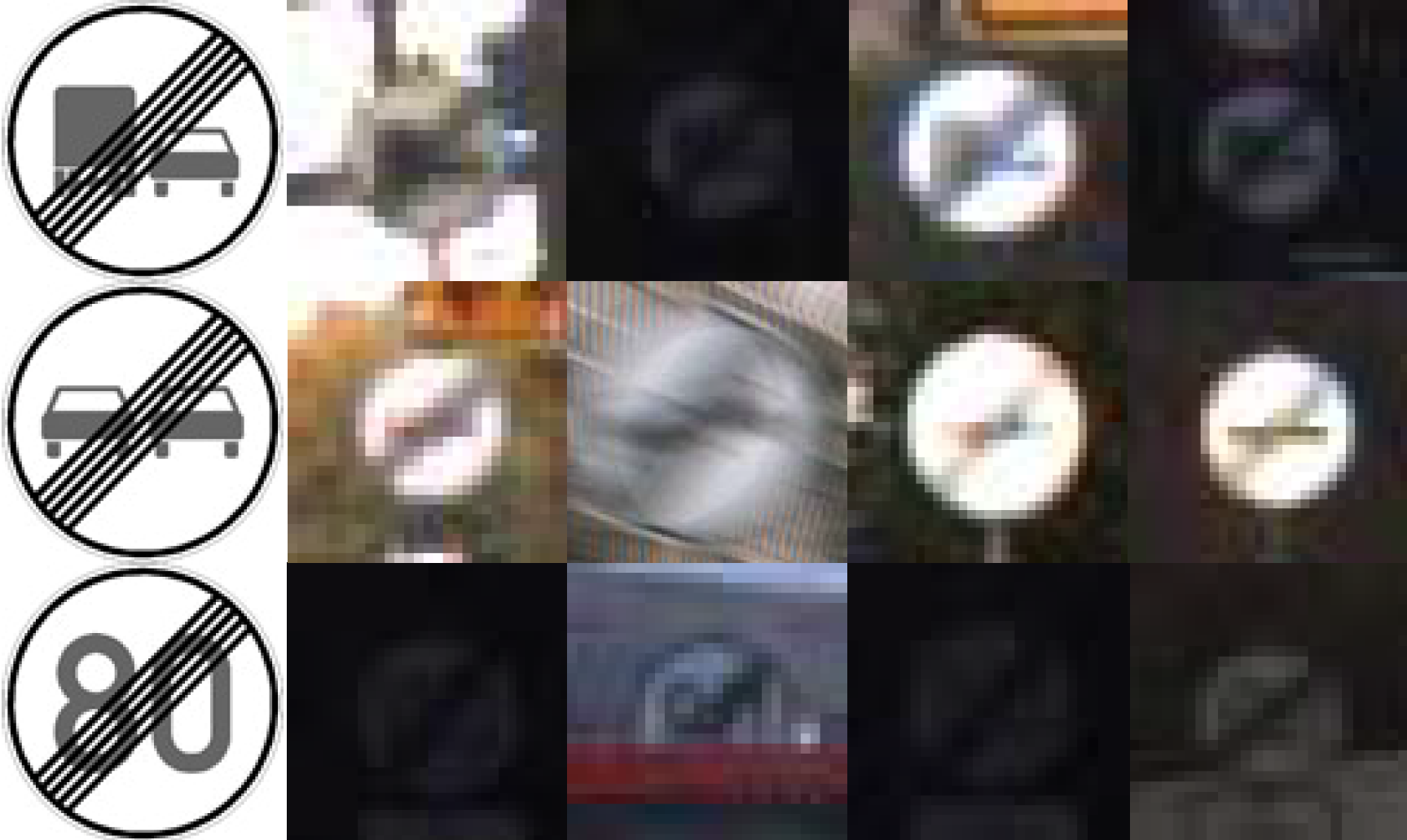}
	\caption{Some of failure cases for unseen recognition.}
	\label{fig:failurecase}
\end{figure}

The proposed \emph{Quadruplet} network has two main factors: the embedding dimension $D$ and the number of pairwise loss terms.
We evaluate these on the validation sets $\Psi_s$ and $\Psi_u$ of GTSRB-all. 
The \emph{Quadruplet} network is trained only with the seen training set $\Phi _s$.

\emph{Embedding dimension}:
We conduct the evaluation while varying the embedding dimension $D$, as reported in \Tref{tab:selfeval_dim}. 
We vary $D$ from $50$ to $150$ and measure the one-shot classification accuracy. 
We observed that the overall average accuracy across the seen and unseen classes peaked at $D{=}100$. 
While the unseen performance may increase further beyond $D=150$, since this sacrifices the seen performance, we select the dimension with the best average accuracy, \ie, $D=100$, as the reference henceforth.


\emph{pairwise loss terms}:
The quadruplet model gives 6 possible pairwise distances between outputs. Intuitively, three possible options, \ie, \textsf{HingeM-3}, \textsf{HingeM-5} and \textsf{HingeM-6} in
Eqs.~(\ref{eqn:loss3}-\ref{eqn:loss6}), satisfy our co-domain embedding property. 
The results in \Tref{tab:selfeval_loss} show a trade-off between different losses. \textsf{HingeM-3} performs worse on both seen and unseen classes than the others, while the other two have a clear trade-off. This implies that \textsf{HingeM-3} is not producing enough information (\ie, gradients) to learn a good feature space. \textsf{HingeM-5} outperforms Loss6 on unseen classes but \textsf{HingeM-6} is better on seen classes. We suspect that complex pairwise relationships among real samples may lead to a feature space which is overly adapted to seen classes.
This trade off can be used to adjust between improving the  seen class performance or regularizing the model for greater flexibility.
We report the following results based on \textsf{HingeM-5} hereafter.

\begin{table}
	\centering
	
	\caption{One-shot classification (Top 1-NN) accuracy ($\%$) on the unseen classes. Performance on seen classes are shown as an additional reference.}
	\label{tbl:top1_nn}
	
	\resizebox{1\linewidth}{!}{%
		\begin{tabular}{cccccccccc}
			\toprule
			\multirow{3}{*}{Network type} & & \multicolumn{6}{c}{Datasets} \\ \cmidrule{3-10}
			& Anchor & \multicolumn{2}{c}{GTSRB} & \multicolumn{2}{c}{GTSRB subset} & \multicolumn{2}{c}{TT100K}\\ 
			\cmidrule(l{2pt}r{2pt}){3-4} \cmidrule(l{2pt}r{2pt}){5-6} \cmidrule(l{2pt}r{2pt}){7-8} \cmidrule(l{2pt}r{0pt}){9-10} 
			& & Seen & Unseen & Seen & Unseen & Seen & Unseen \\ 
			\midrule
			IdsiaNet & $\times$ &  74.6 & 40.2 & 59.4 & 34.7 & 87.7 & 14.6 \\ 
			Triplet & Real &  62.6 & 33.9 & 54.9 & 23.5 & 5.9 & 1.0 \\ \midrule 
			Triplet-DA & Template &  86.0 & \textbf{54.5} & 60.8 & 47.6 & 86.2 & 67.1 \\ 
			\midrule
			Quad.+\textsf{contrastive-5} & Template &  \textbf{92.4} & 42.7 & 69.2 & 41.8 & 92.0 & \textbf{67.5} \\ 
			Quad.+\textsf{hingeM} & Template &  90.8 & 45.2 & \textbf{70.1} & \textbf{47.8} & \textbf{94.9} & 67.3 \\ 
			\bottomrule
		\end{tabular}
	}

	\caption{One-shot classification (Top 1-NN) accuracy ($\%$) from GTSRB-all to TT100K-c.}
	\label{table:transfer}
	
	\resizebox{0.7\linewidth}{!}{%
		\begin{tabular}{cc}
			\toprule
			\multirow{2}{*}{\begin{tabular}[c]{@{}c@{}} Network trained on GTSRB  \end{tabular}} & Top1 NN \\  & sample average 
			\\ 	\midrule
			IdsiaNet            & 36.5 			 \\ 
			Triplet DA		    & 34.1 			 \\ 
			Quadruple+\textsf{hingeM}    & \textbf{42.3} 			 \\ 
			\bottomrule
		\end{tabular}
	}
\end{table}

\paragraph{Comparison to other methods}
We compare the results here with those of other methods on the test sets $\Omega_s$ and $\Omega_u$ of three datasets. 
All networks are trained only with the seen training and validation set  ${\Phi _s} \cup {\Psi _s}$ of respective datasets.

For \emph{IdsiaNet}, we use the activation of \texttt{FC1} for feature embedding.
For the other networks, we use the final embedding vectors.
For the \emph{Quadruplet} network, we test two cases with \textsf{contrastive-5} and \textsf{HingeM-5}.

\Tref{tbl:top1_nn} shows the results on the three datasets, where \emph{Triplet} has the lowest performance, while \emph{Triplet-DA} performs well. Moreover, our quadruplet network outperforms \emph{Triplet-DA} in most of cases for both seen and unseen classes. This supports our hypothesis that the template (also a different domain) anchor based metric learning and the quadruplet relationship may be helpful for generalization purposes.

To check the embedding capability of each approach further, we trained each model on GTSRB ${\Phi _s} \cup {\Psi _s}$ and tested on TT100K-c ${\Omega_s} \cup {\Omega_u}$. This experiment qualifies how the networks perform on two completely different traffic sign datasets. \Tref{table:transfer} shows the top1-NN performance of each model. Our model performs best tn the transfer scenario, which implies that good feature representation is learned.

In order to demonstrate quantitatively the behavior of the \emph{Quadruplet} network, we visualize unseen examples that are often confused by \emph{Quadruplet} in \Fref{fig:failurecase}. 
The unseen classes of examples that are highly similar to other classes are challenging even for humans; furthermore due to the poor illumination condition, motion blur and low resolution.


\subsection{Learned representation evaluation}

In this experiment, we evaluate the generalization behavior of the proposed quadruple network, which is analyzed by comparing the representation power of each network over unseen regimes (and seen class cases as a reference).
In order to assess the representation quality of each method purely, we pre-train competing models and our model for ${\Phi _s} \cup {\Psi _s}$ of each dataset (\ie, only on seen classes), fix the weights of these models, and use the activations of \texttt{FC1} (350 dimensions) of them as a feature.
Given the features extracted by each model, we measure the representation performance by separately training the standard multi-class SVM~\cite{chang2011libsvm} with the radial basis kernel such that the performance heavily relies on feature representation itself.
\footnote{We follow an evaluation method conducted in \cite{tran2015learning}, where the qualities of deep feature representations are evaluated in the same way.}

We use identical SVM parameters (nonlinear RBF kernel, $C=100, tol=0.001$) in all experiments.
In contrast to \texttt{FC1} trained only on seen data, the SVM model is trained on both \emph{seen} and \emph{unseen} classes with an equal number of instances per class, and we vary the number of per class training samples: 10, 50, 100, and 200. SVM training samples are randomly sampled from the set ${\Phi _s} \cup {\Psi _s} \cup {\Phi _u} \cup {\Psi _u}$ (\Fref{fig:datapartition}), and the entire test set ${\Omega_s} \cup {\Omega_u}$ is used for the evaluation. We report the average score and confidence interval by repeating the experiments 100 times for the case of [No. instances$/$class: 10] and 10 times for the cases of [No. instances$/$class: 50, 100, 200]. For each sampling, we fix a random seed and test each model with the same data for a fair comparison.

With the three datasets, we show the results of the unseen ${\Omega_u}$ and seen ${\Omega_s}$ test cases in Table~\ref{tbl:svm} and \ref{tbl:svm_seen}, respectively.
The unseen and seen class datasets are mutually exclusive; hence, the errors should be compared in each dataset independently, \eg, the numbers in \Tref{tbl:svm}-(a) are not directly comparable with those in \Tref{tbl:svm_seen}-(a).
Instead, we can observe the algorithmic behavior difference by comparing Table~\ref{tbl:svm} and \ref{tbl:svm_seen} with respect to the relative performance.


\begin{table}[t]
	\centering
	\caption{Feature representation quality comparison for \textbf{unseen classes}.  SVM classification errors ($\%$) are reported with the datasets, GTSRB-all, GTSRB-sub and TT100K-c, and according to the number of SVM training instances per class. 
Notice that, for all the networks, the classes evaluated in this experiment are not used for training the networks, \ie, \textbf{unseen classes} are used only for SVM training.
Marked in bold are the best results for each scenario, as well as other results with an overlapping confidence interval with 95$\%$.
}
	\label{tbl:svm}
	
	\begin{subtable}[h]{1\linewidth}
		\centering
		\resizebox{1\linewidth}{!}{%
			\begin{tabular}{ccccccccccc}
				\toprule
				\multirow{2}{*}{Network} & \multicolumn{4}{c}{No. instances$/$class} \\ \cmidrule{2-5}
				& 200 & 100 & 50 & 10 \\ 
				\midrule
				IdsiaNet
				& 6.14 ($\pm$0.50) & 6.82 ($\pm$0.57) & 7.82 ($\pm$0.64) & 11.01 ($\pm$0.31) \\
				Triplet  & 7.22 ($\pm$0.33) & 7.94 ($\pm$0.33) & 8.79 ($\pm$0.40) & 15.83 ($\pm$0.38) \\
				Triplet-DA  & 9.04 ($\pm$0.30) & 9.35 ($\pm$0.42) & 10.26 ($\pm$0.63) & 15.17 ($\pm$0.29) \\
				\midrule
				Quad.+\textsf{cont.} & 5.30 ($\pm$0.14) & 6.09 ($\pm$0.39) & 6.96 ($\pm$0.37) & 9.59 ($\pm$0.23) \\
				Quad.+\textsf{hingeM} & \textbf{3.77 ($\pm$0.29)} & \textbf{3.86 ($\pm$0.27)} & \textbf{4.13 ($\pm$0.30)} & \textbf{7.69 ($\pm$0.28)} \\
				\bottomrule
			\end{tabular}
		}
		\caption{GTSRB-all.}
		\label{tbl:gtsrb_all}
	\end{subtable}

	\begin{subtable}[h]{1\linewidth}
		\centering
		\resizebox{1\linewidth}{!}{%
			\begin{tabular}{ccccccccccc}
				\toprule
				\multirow{2}{*}{Network} & \multicolumn{4}{c}{No. instances$/$class} \\ \cmidrule{2-5}
				& 200 & 100 & 50 & 10 \\
				\midrule
				IdsiaNet
				& 17.89 ($\pm$0.41) & 18.28 ($\pm$0.60) & 18.34 ($\pm$0.79) & 25.84 ($\pm$1.52) \\
				Triplet & 17.39 ($\pm$0.67) & 18.22 ($\pm$0.92) & 20.49 ($\pm$1.64) & 30.53 ($\pm$2.47) \\
				Triplet-DA  & 15.84 ($\pm$0.38) & 16.77 ($\pm$0.73) & 18.24 ($\pm$0.48) & 28.43 ($\pm$1.61) \\ 
				\midrule
				Quad.+\textsf{cont.}  & \textbf{12.12 ($\pm$0.33)} & \textbf{11.72 ($\pm$0.40)} & \textbf{12.83 ($\pm$0.43)} & \textbf{18.83 ($\pm$1.35)} \\
				Quad.+\textsf{hingeM}  & 13.81 ($\pm$0.31) & 14.26 ($\pm$0.38) & 15.59 ($\pm$0.52) & 24.31 ($\pm$0.45) \\
				\bottomrule
			\end{tabular}
		}
		\caption{GTSRB-sub.}
		\label{tbl:gtsrb_sub}
	\end{subtable}
    
	\begin{subtable}[h]{0.8\linewidth}
		\centering
		\resizebox{1\linewidth}{!}{%
			\begin{tabular}{ccccccccccc}
				\toprule
				\multirow{2}{*}{Network} & \multicolumn{1}{c}{No. instances$/$class} \\ \cmidrule{2-2}
				& 20 \\
				\midrule
				IdsiaNet~\cite{cirecsan2012multi}  
				& 3.71 ($\pm$0.35) \\
				Triplet~\cite{hoffer2015deep}  & 3.70 ($\pm$0.30) \\ 
				Triplet-DA  & 5.05 ($\pm$0.45) \\ 
				\midrule
				Quad.+\textsf{cont.}  & \textbf{2.97 ($\pm$0.31)} \\
				Quad.+\textsf{hingeM}  & \textbf{2.87 ($\pm$0.24)} \\
				\bottomrule
			\end{tabular}
		}
		\caption{TT100k-c.}
		\label{tt100k_all}
	\end{subtable}
\end{table}


\begin{table}[t]
	\centering
	\caption{Feature representation quality comparison for \textbf{seen classes}. 
SVM classification errors ($\%$) are reported with the datasets, GTSRB-all, GTSRB-sub and TT100K-c, and according to the number of SVM training instances per class. 
Marked in bold are the best results for each scenario, as well as other results with an overlapping confidence interval with 95$\%$.
}
	\label{tbl:svm_seen}
	
	\begin{subtable}[h]{1\linewidth}
		\centering
		\resizebox{1\linewidth}{!}{%
			\begin{tabular}{ccccccccccc}
			\toprule
			\multirow{2}{*}{Network type} & \multicolumn{4}{c}{No. instances$/$class} \\ \cmidrule{2-5}
			& 200 & 100 & 50 & 10 \\ 
			\midrule
			IdsiaNet & \textbf{3.70 ($\pm$0.09)} & \textbf{4.24 ($\pm$0.14)} & \textbf{4.72 ($\pm$0.18)} & 6.52 ($\pm$0.10) \\
			Triplet & 5.00 ($\pm$0.08) & 5.51 ($\pm$0.12) & 6.22 ($\pm$0.16) & 9.11 ($\pm$0.15) \\
			Triplet-DA & 4.75 ($\pm$0.12) & 5.30 ($\pm$0.10) & 6.08 ($\pm$0.22) & 8.99 ($\pm$0.13) \\
			\midrule
			Quad.+\textsf{cont.} & 4.49 ($\pm$0.09) & \textbf{4.50 ($\pm$0.12)} & \textbf{4.65 ($\pm$0.11)} & \textbf{5.61 ($\pm$0.13)} \\
			Quad.+\textsf{hingeM} & 4.46 ($\pm$0.08) & 4.63 ($\pm$0.12) & \textbf{4.81 ($\pm$0.15)} & 5.98 ($\pm$0.08) \\
			\bottomrule
		\end{tabular}
		}
		\caption{GTSRB-all.}
		\label{tbl:gtsrb_all_seen}
	\end{subtable}

	\begin{subtable}[h]{1\linewidth}
		\centering
		\resizebox{1\linewidth}{!}{%
			\begin{tabular}{ccccccccccc}
			\toprule
			\multirow{2}{*}{Network type} & \multicolumn{4}{c}{No. instances$/$class} \\ \cmidrule{2-5}
			& 200 & 100 & 50 & 10 \\
			\midrule
			IdsiaNet & 7.55 ($\pm$0.25) & 8.64 ($\pm$0.22) & 10.13 ($\pm$0.43) & 17.19 ($\pm$0.23) \\
			Triplet & 10.88 ($\pm$0.25) & 12.35 ($\pm$0.31) & 14.28 ($\pm$0.27) & 24.50 ($\pm$0.30) \\
			Triplet-DA & 8.78 ($\pm$0.30) & 10.44 ($\pm$0.2) & 12.80 ($\pm$0.37) & 21.15 ($\pm$0.28) \\ 
			\midrule
			Quad.+\textsf{cont.} & \textbf{6.04 ($\pm$0.16)} & \textbf{6.88 ($\pm$0.11)} & \textbf{7.92 ($\pm$0.28)} & \textbf{12.51 ($\pm$0.21)} \\
            Quad.+\textsf{hingeM} & 6.57 ($\pm$0.15) & 7.54 ($\pm$0.23) & 8.87 ($\pm$0.31) & 14.02 ($\pm$0.20) \\
			\bottomrule
		\end{tabular}
		}
		\caption{GTSRB-sub.}
		\label{tbl:gtsrb_sub_seen}
	\end{subtable}
    
	\begin{subtable}[h]{0.8\linewidth}
		\centering
		\resizebox{1\linewidth}{!}{%
			\begin{tabular}{ccccccccccc}
			\toprule
			\multirow{2}{*}{Network type} & \multicolumn{1}{c}{No. instances$/$class} \\ \cmidrule{2-2}
			& 20 \\
			\midrule
			IdsiaNet~\cite{cirecsan2012multi} & \textbf{4.23 ($\pm$0.22)} \\
			Triplet~\cite{hoffer2015deep} & 5.30 ($\pm$0.14) \\ 
			Triplet-DA (100) & 5.53 ($\pm$0.17) \\ 
			\midrule
			Quad.+\textsf{cont.} & 5.39 ($\pm$0.23) \\
			Quad.+\textsf{hingeM} & \textbf{4.33 ($\pm$0.14)} \\
			\bottomrule
		\end{tabular}
		}
		\caption{TT100k-c.}
		\label{tbl:tt100k_all_seen}
	\end{subtable}
\end{table}

In the unseen case in \Tref{tbl:svm}, for all of the results, the proposed \emph{Quadruplet} variants outperform the other competing methods, supporting the fact that our method generalizes well in limited sample regimes by virtue of the richer information from the quadruples.
As an addendum, interestingly, \emph{Triplet-DA} performs better than \emph{Triplet} only with GTSRB-sub. We postulate that, in terms of feature description with some amount of strong supervision support, \emph{Triplet} generalizes better than \emph{Triplet-DA}, due
inherently to the number of possible triplet combinations of \emph{Triplet-DA} ($|\mathcal{T}|{\cdot}|\mathcal{X}|^2$, and generally $|\mathcal{T}| {\ll} |\mathcal{X}|$) is much smaller than that of \emph{Triplet} due to the use of a template anchor as an element of triplet, while \emph{Triplet} improves all the pairwise possibility of real data (\ie, $|\mathcal{X}|^3$). 
For the quadruplet case, more pairwise relationships results in higher number of tuple combinations and leads to better generalization, even with the use of templates.



In the seen class case in \Tref{tbl:svm_seen}, our method and \emph{IdsiaNet} show comparable performance outcomes with the best accuracy levels, while \emph{Triplet} and \emph{Triplet-DA} do not perform as well. 
We found that the proposed \emph{Quadruplet} performs better than the other approaches when the number of training instances is very low, \ie, 10 samples per class. 
More specifically, \emph{IdsiaNet} performs well in the seen class case, but not in the unseen class case compared to \emph{Quadruplet}.
This indicates that the embedding of IdsiaNet is trapped in seen classes rather than in general traffic sign appearances.
On the other hand, the proposed method learns more general representation in that its performance in both cases is higher than those of its counterparts.
We believe that this is due to the regularization effect caused by the usage of templates.

\section{Conclusion}

In this study, we have proposed a deep quadruplet for one-shot learning and demonstrated its  performance on the unseen traffic-sign recognition problem with template signs.
The idea is that by composing a quadruplet with a template and real examples, the combinatorial relationships enable not only domain adaptation via co-embedding learning, but also generalization to the unseen regime.
Because the proposed model is very simple, it can be extended to other domain applications, where any representative anchor can be given.
We think that $N>4$-tuple generalization is interesting as a future direction in that there must be a trade-off between over-fitting and memory capacities.

\section*{Acknowledgment}
This work was supported by DMC R\&D Center of Samsung Electronics Co.

{
\bibliographystyle{aaai}
\bibliography{egbib}
}

\cleardoublepage
\section*{Supplementary Materials}
Here, we present additional details of the baseline network and datasets that could not be included in the main text due to space constraints. All figures and references in this supplementary file are consistent with the main paper.

\section{Base neural network architecture}

\paragraph{Base network}
We select the network based on classification performance under the assumption that high classification performance stems from good feature embedding capabilities. 
The Idsia network\cite{cirecsan2012multi} is a winner in a German Traffic Sign Recognition Benchmark (GTSRB) competition. It was reported that variants of the Idsia network perform even better. The variant IdsiaNet uses a deeper convolution kernel size for each layer (\Tref{tbl:idsianet-layers}). We use the best performing Idsia-like network~\footnote{IdsiaNet~\cite{cirecsan2012multi} is tuned to achieve better performance. Please check the repository, \url{https://github.com/moodstocks/gtsrb.torch\#gtsrbtorch}} for all three embedding schemes in the experiments. 


\begin{table}[h]
	\centering
	\caption{Variant IdsiaNet structure. Zero pad 2 before each conv layers. (LCN: local contrast normalization)}
	\label{tbl:idsianet-layers}
	\begin{tabular}{|c|c|c|c|}
		\hline
		Layer & Type & Size & kernel size \\ \hline
		0	& input		& 48x48x3		& - \\ \hline
		1	& conv1		& 46x46x150	& 7x7 \\
		2	& ReLU		& 46x46x150	& - \\
		3	& Max pooling	& 23x23x150	& 2x2 \\
		4	& LCN		& 23x23x150	& 7x7 \\ \hline
		5	& conv2		& 24x24x200	& 4x4 \\
		6	& ReLU		& 24x24x200	& - \\
		7	& Max pooling	& 12x12x200	& 2x2 \\
		8	& LCN		& 12x12x200	& 7x7 \\ \hline
		9	& conv3		& 13x13x300	& 4x4 \\
		10	& ReLU		& 13x13x300	& - \\
		11	& Max pooling	& 6x6x300	& 2x2 \\
		12	& LCN		& 6x6x300	& 6x6 \\ \hline
		13	& fc1		& 		350		& 10800 $\to$ 350 \\
		14	& ReLU		& 		350		& - \\
		15	& fc2		& 		43		& 350 $\to$ 43 \\ 
		16	& softmax	& 		43		& - \\ 
		\hline
	\end{tabular}
\end{table}

\section{Details of datasets}

\begin{figure}[p]
	\begin{center}
		{\includegraphics[width=1.0\linewidth]{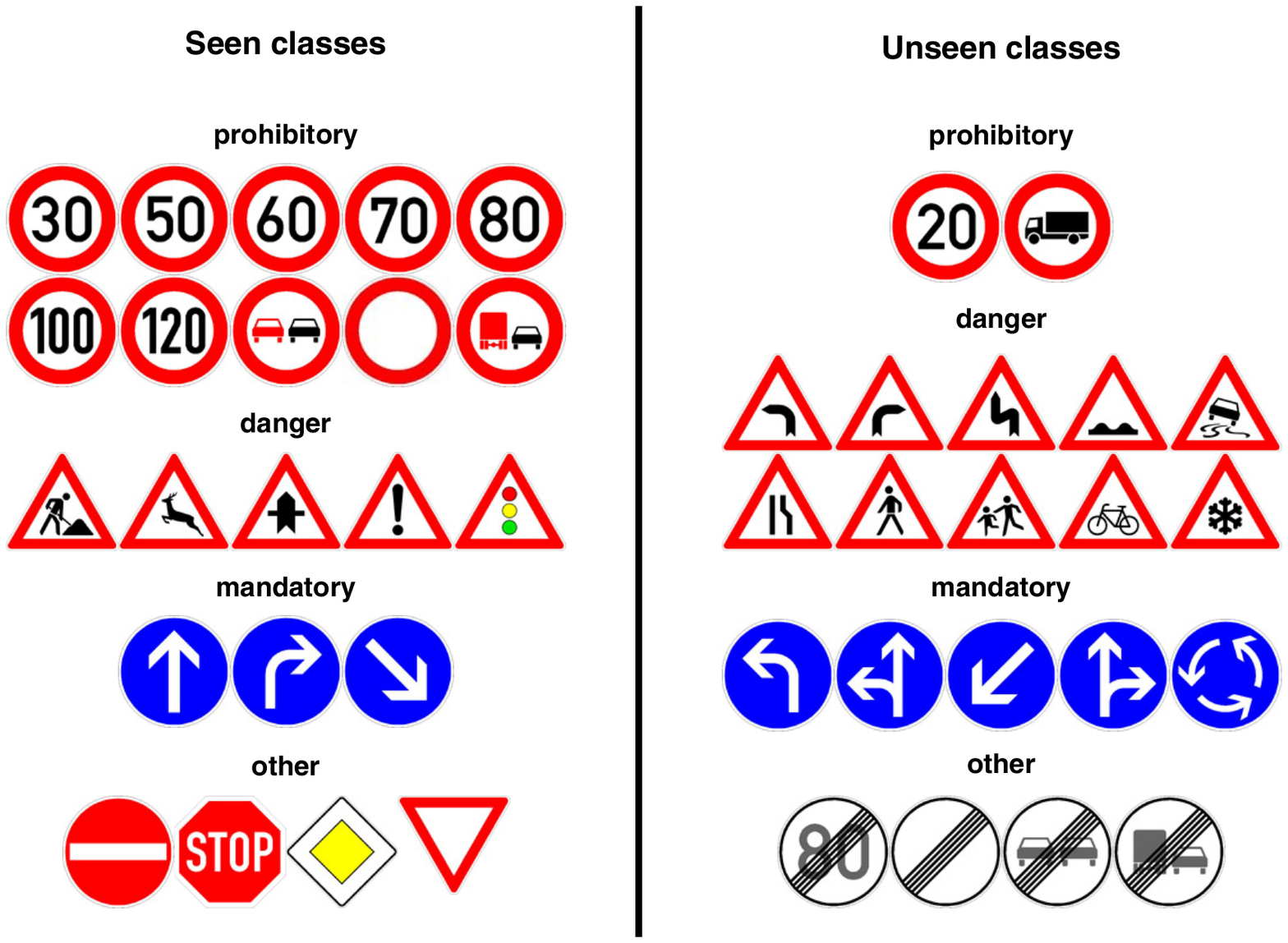}}\hspace{2mm}
	\end{center}
	\vspace{-1mm}
	\caption{GTSRB template images.}
	\label{fig:GTSRB_template}
	\vspace{-1mm}	
\end{figure}

\begin{figure}[p]
	\begin{center}
		{\includegraphics[width=1.0\linewidth]{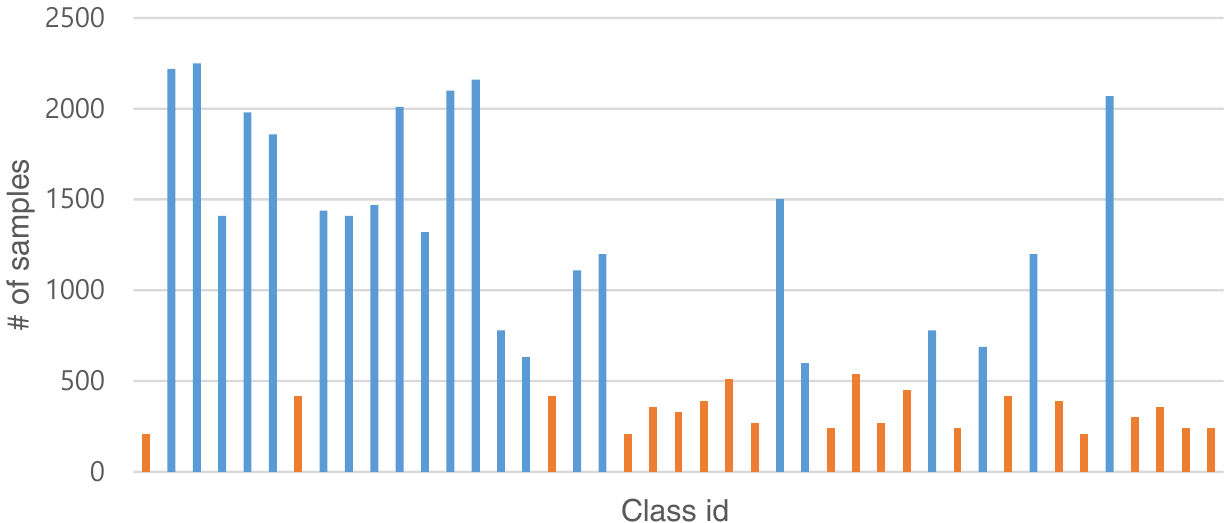}}\hspace{2mm}
	\end{center}
	\vspace{-1mm}
	\caption{GTSRB class sample distribution. \textcolor{blue}{Blue}: seen classes (\#22). \textcolor{orange}{Orange}: unseen classes (\#21).}
	\label{fig:GTSRB_class}
	\vspace{-1mm}
\end{figure}

\begin{figure}[p]
	\begin{center}
		{\includegraphics[width=1.0\linewidth]{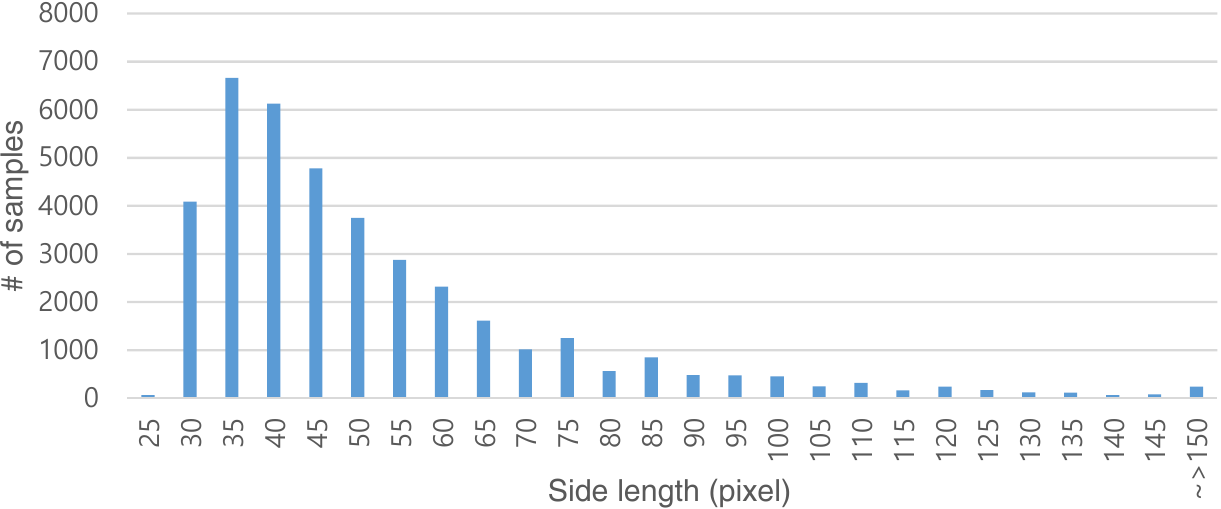}}\hspace{2mm}
	\end{center}
	\vspace{-1mm}
	\caption{GTSRB size distribution. $Side length=(height+width)/2$}
	\label{fig:GTSRB_size}
	\vspace{-1mm}
\end{figure}

\paragraph{GTSRB-all}~{GTSRB~\cite{stallkamp2012man} has been the most widely used and largest dataset for traffic-sign recognition. It contains 43 classes categorized into three super categories: \emph{prohibitory}, \emph{danger} and \emph{mandatory}. The dataset contains illumination variations, blur, partial shadings and low-resolution images. The benchmark is partitioned into \emph{training} and \emph{test} sets. The training set contains 39,209 images, with 30 consecutively captured images grouped, which is referred to here at a track. The test set contains 12,630 images without continuity and thus does not form tracks. The number of samples per class varies, and the sample size also has a wide range (\Fref{fig:GTSRB_class} \& \Fref{fig:GTSRB_size}). 
	To validate the embedding capability of the proposed model, we split the training set into two sets: seen and unseen (\Fref{fig:GTSRB_template}). Our motivation is to overcome the data imbalance caused by rare classes. We selected 21 classes that have the fewest samples as the unseen class, using the remaining 22 classes as the seen class. For training, we only used seen classes in the training set.}

\paragraph{GTSRB-sub}~{
	To analyze the performance on a dataset with a limited size, we created a subset of GTSRB-all, forming a smaller size, as a class balanced training set. For the training set, we randomly selected seven tracks for each of the seen classes. For the test set, we used the same set as used in GTSRB-all; hence results will be compatible with and comparable to those of GTSRB-all.}

\begin{figure}[p]
	\begin{center}
		{\includegraphics[width=1.0\linewidth]{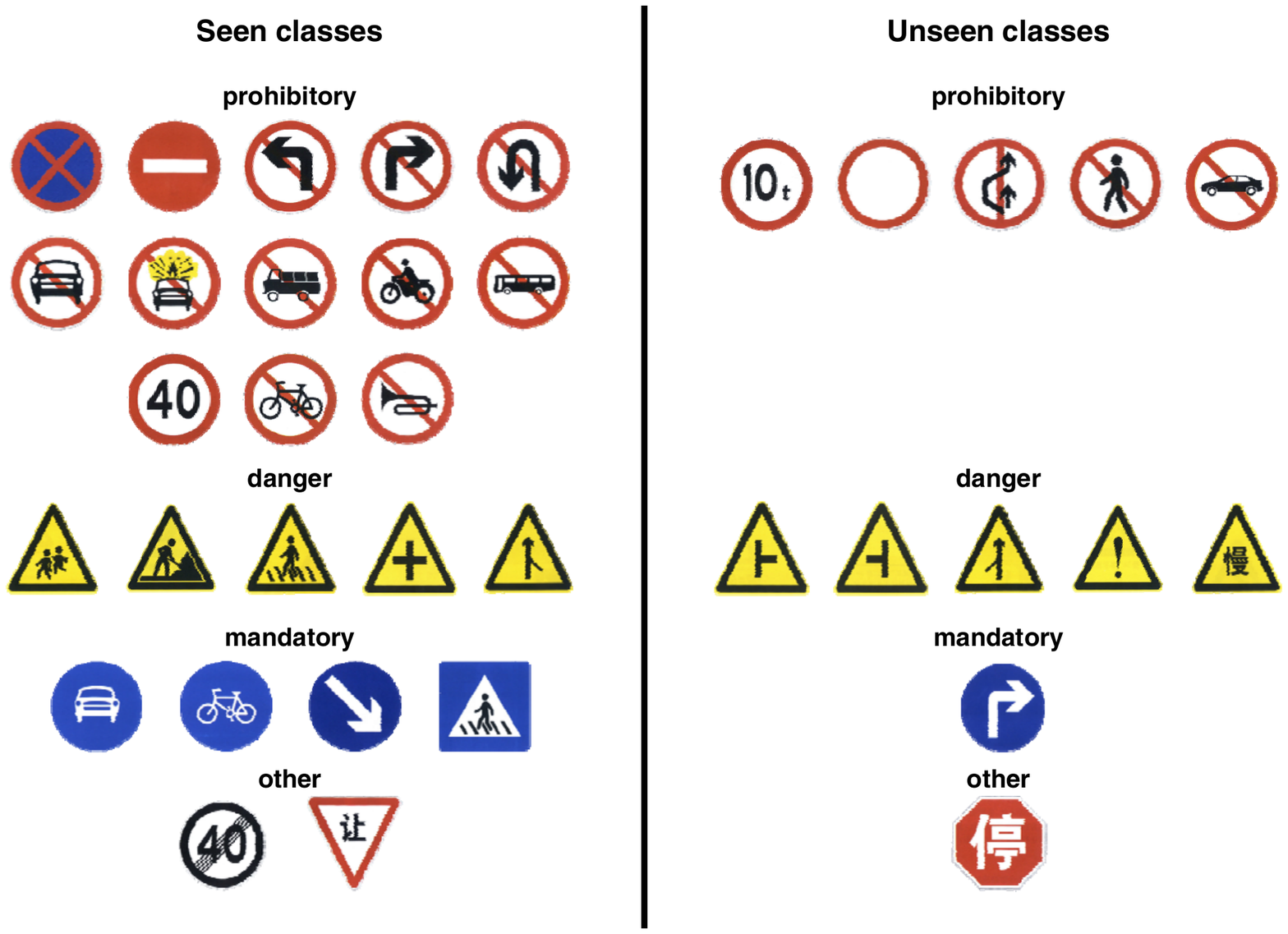}}\hspace{2mm}
	\end{center}
	\vspace{-1mm}
	\caption{TT100K template images.}
	\label{fig:TT100K_template}
	\vspace{-1mm}
\end{figure}

\begin{figure}[p]
	\begin{center}
		{\includegraphics[width=1.0\linewidth]{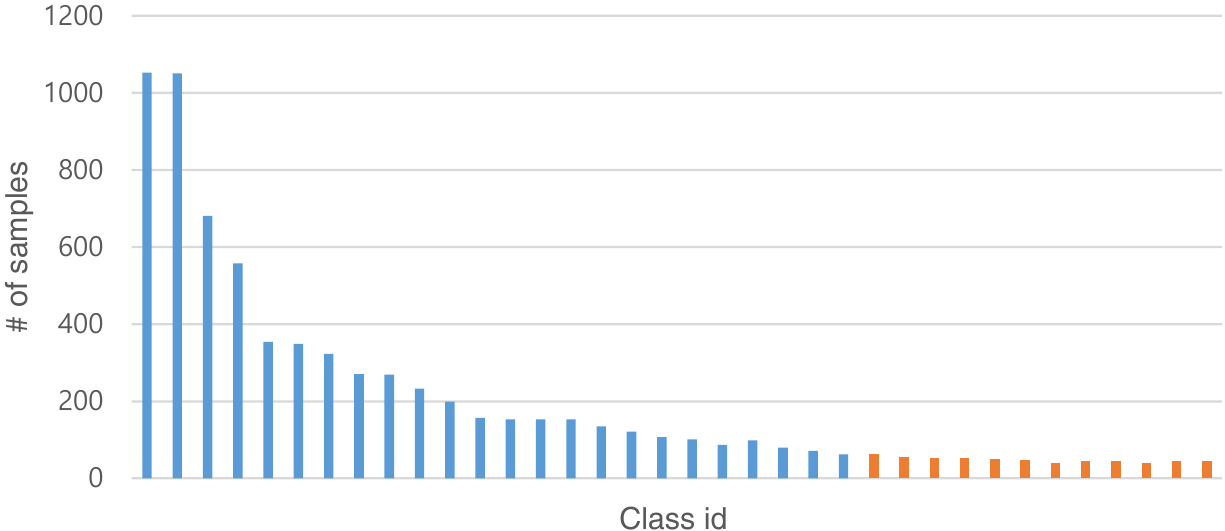}}\hspace{2mm}
	\end{center}
	\vspace{-1mm}
	\caption{TT100K class sample distribution. \textcolor{blue}{Blue}: seen classes (\#24). \textcolor{orange}{Orange}: unseen classes (\#12).}
	\label{fig:TT100K_class}
	\vspace{-1mm}
\end{figure}

\begin{figure}[p]
	\begin{center}
		{\includegraphics[width=1.0\linewidth]{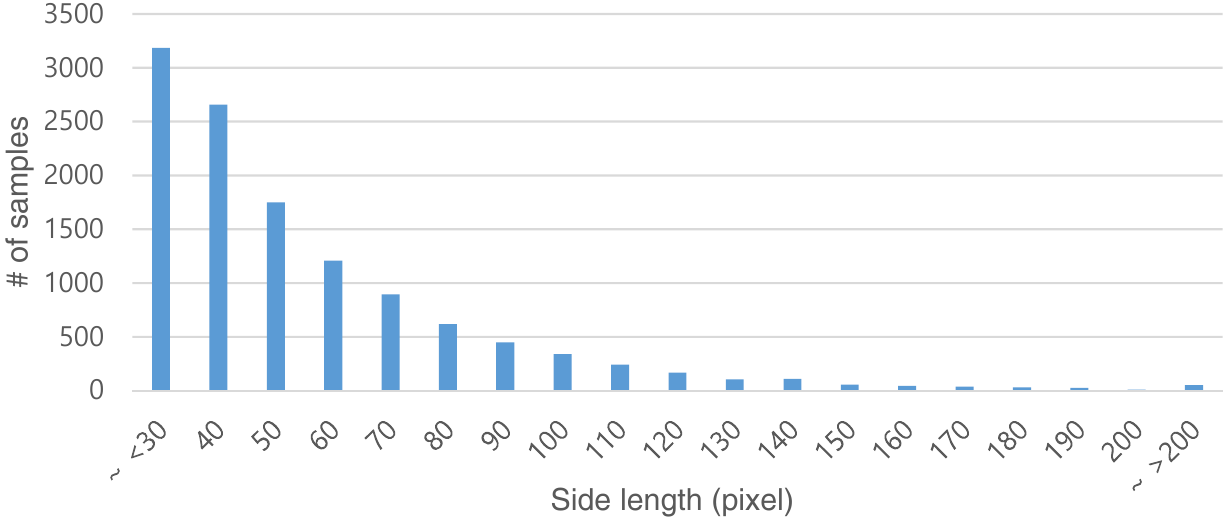}}\hspace{2mm}
	\end{center}
	\vspace{-1mm}
	\caption{TT100K size distribution. $Side length=(height+width)/2$}
	\label{fig:TT100K_size}
	\vspace{-3mm}
\end{figure}

\paragraph{TT100K-c}~{ Tsinghua-Tencent 100K~\cite{zhu2016traffic} (TT100K) is a chinese traffic sign detection dataset that includes more than 200 sign classes. 
	We cropped traffic sign instances from scenes to build a classification dataset (called TT100K-c) to validate our model. Although it contains numerous classes, most of those in each class do not have enough instances to allow the experiment to be conducted. 
	We only selected classes with official templates\footnote{The official templates are provided by  Beijing Traffic Management Bureau,
		\url{http://www.bjjtgl.gov.cn/english/traffic signs/index.html}.} available and a sufficient number of instances. We filtered out instances that have side lengths of less than $20$ pixels, as they are unrecognizable.
	We split TT100K-c into the training and test set according to the number of instances. We set 24 classes that have more than 100 instances as the seen classes, and assign 12 classes with between 50 and 100 instances as the unseen class (\Fref{fig:TT100K_template}). The \emph{training} set includes half of the seen class samples, with the other half going to the test set. 
	All of the unseen class samples belong to the \emph{test} set. The size distribution and the number of samples per class are provided in \Fref{fig:TT100K_class} and \Fref{fig:TT100K_size}, respectively.
}

\end{document}